\documentclass[journal]{IEEEtran}

\usepackage{amsmath,amsfonts,amssymb}
\usepackage{graphicx}
\usepackage[backend=biber,style=numeric-comp,sorting=none]{biblatex}
\usepackage{etoolbox}

\DeclareFieldFormat{labelnumber}{[#1]} 

\makeatletter
\patchcmd{\blx@range@labelnum}{\textendash}{]-[}{}{} 
\patchcmd{\blx@range@labelnum}{--}{]-[}{}{}          
\makeatother

\usepackage{hyperref}
\usepackage{xcolor}
\usepackage{booktabs}
\usepackage{array}
\usepackage{colortbl}
\usepackage{longtable}
\usepackage{subcaption}
\usepackage{url}
\usepackage{multirow}
\usepackage{diagbox}

\AtBeginBibliography{\footnotesize}

\begin{document}

\title{Style-Based Neural Architectures for Real-Time Weather Classification}

\author{Hamed~Ouattara, 
        Pierre~Duthon, 
        Frédéric~Bernardin, 
        Omar~Ait~Aider, 
        and Pascal~Salmane%
\thanks{Corresponding author: Hamed~Ouattara (e-mail: hamed.ouattara@cerema.fr).}}

\markboth{Journal of IEEE Transactions on Neural Networks and Learning Systems,~Vol.~XX, No.~X, Month~Year}%
{Ouattara \MakeLowercase{\textit{et al.}}: Style Transfer Heuristic for Real-Time Detection or Classification of Weather Conditions}

\IEEEpubid{\begin{minipage}{\textwidth}\vspace{1.8cm}\footnotesize
Funded by the European Union (grant no. 101069576). However, the opinions and points of view expressed are solely those of the author(s).\\
Hamed Ouattara, Pierre Duthon, Frédéric Bernardin, and Pascal Salmane are with Cerema, Intelligent Transportation Systems Research Team, France.\\
Omar Ait Aider is with Institut Pascal, France.
\end{minipage}}%
\maketitle

\begin{abstract}
In this paper, we present three neural network architectures designed for real-time classification of weather conditions (sunny, rain, snow, fog) from images. These models, inspired by recent advances in style transfer, aim to capture the stylistic elements present in images. One model, called “Multi-PatchGAN,” is based on PatchGANs used in well-known architectures such as Pix2Pix and CycleGAN, but here adapted with multiple patch sizes for detection tasks. The second model, “Truncated ResNet50,” is a simplified version of ResNet50 retaining only its first nine layers. This truncation, determined by an evolutionary algorithm, facilitates the extraction of high-frequency features essential to capture subtle stylistic details. Finally, we propose “Truncated ResNet50 with Gram Matrix and Attention,” which computes Gram matrices for each layer during training and automatically weights them via an attention mechanism, thus optimizing the extraction of the most relevant stylistic expressions for classification. These last two models outperform the state of the art and demonstrate remarkable generalization capability on several public databases.

Although developed for weather detection, these architectures are also suitable for other appearance-based classification tasks, such as animal species recognition, texture classification, disease detection in medical imaging, or industrial defect identification.
\end{abstract}


\IEEEpeerreviewmaketitle

\section{Introduction}

\IEEEPARstart{R}{eal-time} detection of weather conditions from images is crucial, as weather directly influences human and animal activities. In automated agriculture, irrigation systems, harvesting robots, or spraying drones could benefit from this technology. In the energy sector, where weather affects the production of renewable energies (solar, wind, hydroelectric), real-time weather tracking would make it possible to optimize resource usage based on weather conditions. Given the extensive coverage of cameras (urban surveillance, road networks, satellite imagery), such an algorithm could provide reliable data even in areas without weather stations, thereby enhancing climate models and alert systems. \textbf{Another major advantage of these algorithms lies in their potential use in autonomous vehicles, enabling immediate adaptation of driving to weather conditions.}

\begin{figure}[!t]
    \centering
    \includegraphics[width=0.9\columnwidth]{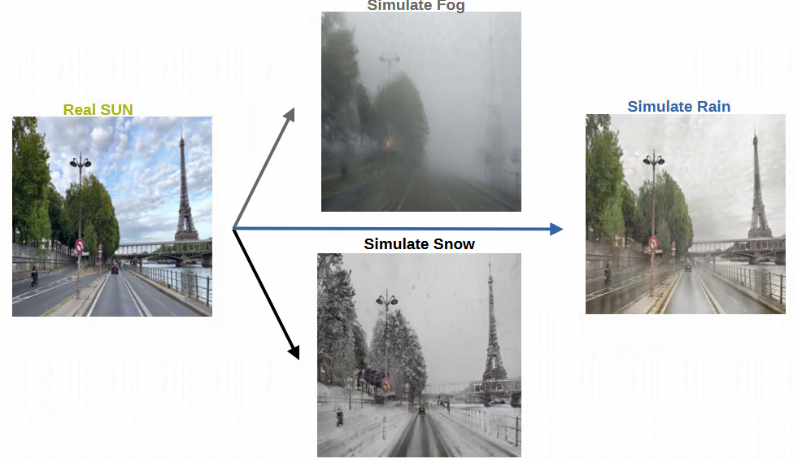}
    \caption{Example of simulating weather conditions using the CycleGAN algorithm.}
    \label{fig:figure1}
\end{figure}

\hypertarget{postulat}{We define the \textbf{style of an image} as the appearance of the objects it contains. This appearance can vary according to several factors such as brightness, shapes, or textures of objects, etc. Thus, in this paper, we hypothesize that weather conditions contribute to the style of an image by influencing the overall appearance of the objects it contains. Indeed, a sunny scene reveals objects in vivid colors, reflecting the broad wavelengths of sunlight. Conversely, during a storm, objects appear duller and more reflective because they are wet. In a foggy scene, object outlines fade, while in a snowy landscape, objects appear uniform due to the whiteness of the snow. In the same way that two painters representing the same scene each paint it in their own style, two different weather conditions give the same scene a distinct appearance specific to each one} (\hyperref[fig:figure1]{Fig. \ref{fig:figure1}}).

Style transfer algorithms enable transferring the appearance of one set of images to another and have made remarkable progress in recent years \cite{key1,key2,key3,gatys2016image,key7,key8,key10,yi2017dualgan,kim2017learning,liu2017unsupervised,huang2018multimodal,choi2018stargan,zhu2017toward,tang2019attention,ulyanov2016instance,huang2017arbitrary,chen2023artfusioncontrollablearbitrarystyle,rombach2022highresolutionimagesynthesislatent,radford2021learningtransferablevisualmodels}. Based on our hypothesis—“weather mainly affects the appearance of objects in an image”—it becomes relevant to explore these algorithms for weather detection and compare them to more traditional methods. Indeed, they can detect and exchange styles between two different image sets, for example transforming photographs into images imitating an artist’s style \cite{gatys2016image}, or simulating various weather conditions \cite{key2}. An example using the CycleGAN model is illustrated in (\hyperref[fig:figure1]{Fig. \ref{fig:figure1}}). Our model codes are available here: \href{https://github.com/Hamedkiri/heuristique_style_transfer_code.git}{\textbf{\textit{code}}}.

\section{Contributions and Paper Organization}
In this paper, we discuss the main contributions \cite{key1, key2, key3, gatys2016image} on style transfer, focusing on how neural network architectures extract stylistic information from an image, particularly weather-related features, to leverage their potential in weather detection. The ability of these architectures to transfer stylistic features from one set of images to another is only useful here for visualizing the information they detect.

We compare three architectures inspired by recent advances in style transfer to achieve \textbf{real-time} automatic detection of weather conditions (rain, snow, fog, sun). The “Truncated ResNet50” and “Truncated ResNet50 + Gram Matrix + Attention” models (\textbf{the latter introduces, for the first time, a learned selection of Gram matrices}) match or exceed the state-of-the-art, displaying strong generalization capabilities, with an F1 score of 98.85\% in tests.

We also demonstrate their potential for other applications, such as disease detection in medical imaging (\hyperlink{otherTasks}{Application of our models to other tasks}). 

\section{Literature Review}

As mentioned in the introduction, we propose using architectures stemming from style transfer algorithms for detection tasks. These approaches will be compared (see Table~\ref{tab:tableau2} in \hyperlink{results}{Section VI}) to standard detection methods, whose literature review is presented in Section III.B.

\subsection{Review of Style Transfer Methods or Heuristics}\label{sec4}

\hypertarget{Gatys}{\subsubsection{Neural Style Transfer and the Work of Leon A. Gatys et al.\ \cite{gatys2016image}}}

\begin{figure}[h]
    \centering
    \includegraphics[width=0.9\columnwidth]{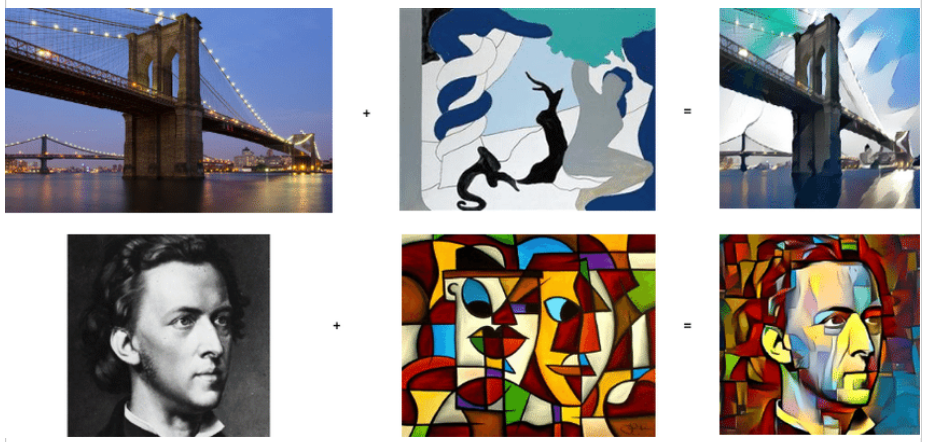}
    \caption{Style transfer using the Leon A. Gatys method\cite{gatys2016image}}
    \label{fig:figure3}
\end{figure}

\begin{figure}[h]
    \centering
    \includegraphics[width=0.9\columnwidth]{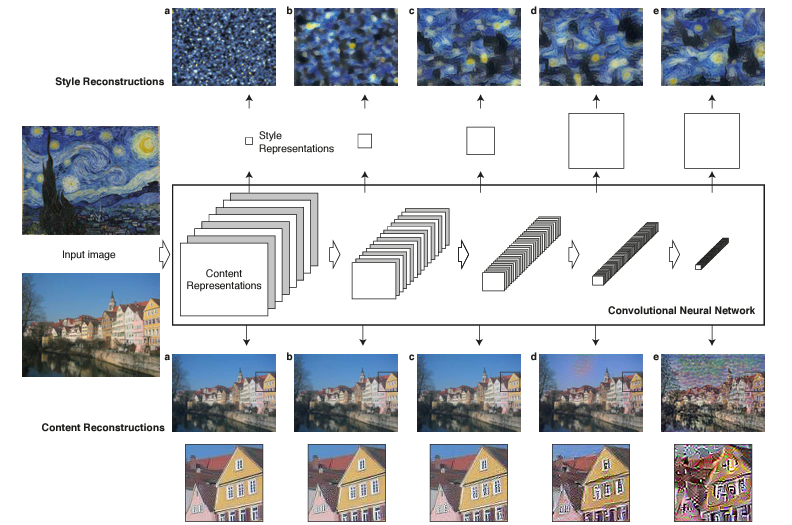}
    \caption{Image from Leon A. Gatys’ article \cite{gatys2016image} showing content and style reconstructions at different layers of a network.}
    \label{fig:figure4}
\end{figure}

Leon A. Gatys et al.\ \cite{gatys2016image}, pioneers in neural style transfer, propose to factorize an image (\hyperref[fig:figure3]{Fig. \ref{fig:figure3}}) by separating its content (objects, animals, etc.) from its style, or the appearance of the objects, which can vary based on factors such as brightness, contrast, or colors. Gatys demonstrates that the high-level layers of a neural network pre-trained for classification or detection tasks effectively capture the presence of objects regardless of their appearance (\hyperref[fig:figure4]{Fig. \ref{fig:figure4}}). Indeed, these networks are designed to robustly recognize objects under different conditions, often by means of data augmentation.

Furthermore, Gatys shows that the style or appearance of objects can be extracted by examining the correlations \cite{gatys2016image} ,\cite{gatys2015texturesynthesisusingconvolutional} among the feature maps of the same layer in a neural network. Each layer may contain several filters, each generating a feature map indicating that filter’s activation across different locations of the image. \hypertarget{Gram_matrice}{The idea is that the stylistic elements defining the appearance of objects can be captured through correlations (or similarities) between these feature maps, reflecting the overall relationships among the patterns present in the image. To obtain these correlations, Gatys proposes computing the \textcolor{red}{Gram matrix}, which corresponds to the dot product of vectorized feature maps in the same layer. It is equivalent to the covariance matrix in statistics.}

\[
\mathbf{G} = \begin{bmatrix}
    c_{1} \cdot c_{1} & c_{1} \cdot c_{2} & c_{1} \cdot c_{3} \\
    c_{2} \cdot c_{1} & c_{2} \cdot c_{2} & c_{2} \cdot c_{3} \\
    c_{3} \cdot c_{1} & c_{3} \cdot c_{2} & c_{3} \cdot c_{3}
\end{bmatrix}
\]

\textbf{Gram matrix:} This matrix represents the correlations between the vectorized feature maps $c_{i}$ of a layer containing three filters. Each entry $c_{i} \cdot c_{j}$ is the dot product between the feature vectors $c_{i}$ and $c_{j}$.

It is important to note that Gram matrices from lower-level layers capture fine stylistic elements, as they are built from filters detecting only local patterns. Conversely, Gram matrices from deeper layers represent more global stylistic elements, given that the \hyperlink{champ_receptif}{receptive field} of their filters is larger (\hyperref[fig:figure6]{Fig. \ref{fig:figure6}}).

In \cite{gatys2016image}, Gatys uses a pre-trained VGG network \cite{simonyan2015deepconvolutionalnetworkslargescale} to separate the content and style of an image.

Considering weather condition detection (since we view weather as part of an image’s style), we can note:

\begin{itemize}
\item \textbf{The stylistic elements of an image can be extracted by examining the correlations among the feature maps in different layers of a neural network trained to recognize patterns in that image.}

\item \textbf{In a deep network trained for detection or classification, the higher-level layers are designed to ignore the appearance of objects in order to recognize them under a wide range of conditions.}
\end{itemize}

\textbf{Important:} Leon A. Gatys et al.\ highlight a major limitation of their method when applied to real photographs or images: using the Gram matrix for style capture does not preserve the spatial coherence of stylistic information \cite{gatys2016image}. This affects the quality of style transfer, particularly between two real images, such as a sunny scene and a rainy scene. In our opinion, this loss of spatial coherence can lead to a loss of information and compromise the relevance of the extracted stylistic data. Hence, it is worth exploring other approaches to better preserve correlations among feature maps, such as the work of Phillip Isola \cite{key1}, who introduced PatchGAN.

\hypertarget{Isola}{\subsubsection{The Pix2Pix Model and the Work of Phillip Isola et al.\ \cite{key1}}}

To preserve the spatial coherence of style elements, Phillip Isola introduced PatchGAN \cite{key1}. Isola uses Generative Adversarial Networks (GANs) \cite{Goodfellow2014} for style transfer between two sets of images having different styles. \hypertarget{patchgan}{We are particularly interested in how stylistic information is extracted, namely PatchGAN.}

\begin{figure}[h]
    \centering
    \includegraphics[width=0.9\columnwidth]{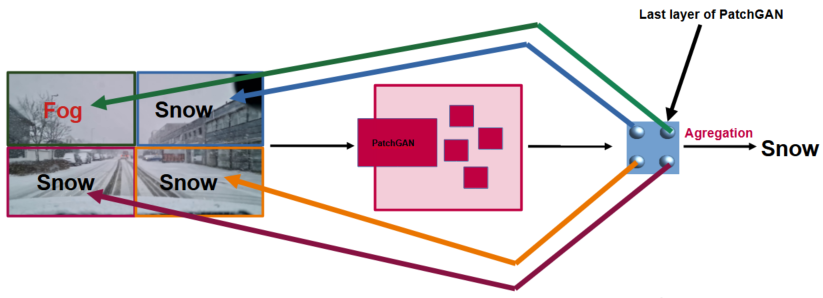}
    \caption{Basic PatchGAN architecture for weather detection}
    \label{fig:figure5}
\end{figure}

\begin{figure}[ht]
    \centering
    \begin{subfigure}{0.8\columnwidth}
        \centering
        \includegraphics[width=\linewidth, height=4cm]{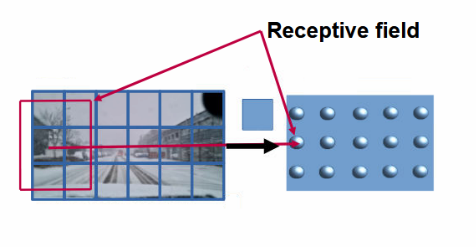}
        \captionsetup{labelformat=empty}
        \caption{First layer}
    \end{subfigure}

    \vspace{0.5cm}

    \begin{subfigure}{0.8\columnwidth}
        \centering
        \includegraphics[width=\linewidth, height=4cm]{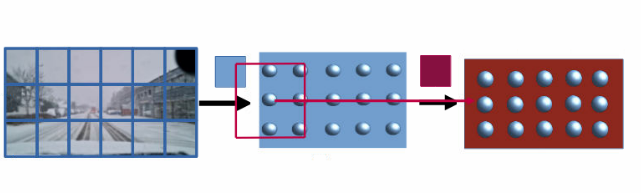}
        \captionsetup{labelformat=empty}
        \caption{Second layer}
    \end{subfigure}

    \vspace{0.5cm}

    \begin{subfigure}{0.8\columnwidth}
        \centering
        \includegraphics[width=\linewidth, height=4cm]{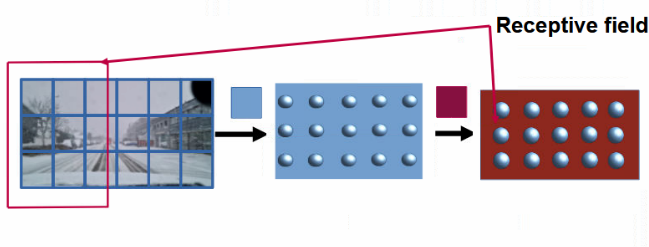}
        \captionsetup{labelformat=empty}
        \caption{Third layer}
    \end{subfigure}
    
    \caption{The receptive field of neurons grows with the number of layers, creating a bottleneck effect for information among neurons.}
    \label{fig:figure6}
\end{figure}

\hypertarget{champ_receptif}{The receptive field of a neuron is the region of the input image that directly or indirectly influences that neuron’s activation.} The uniqueness of PatchGAN lies in how the network perceives the image. The neurons in the final layer have disjoint receptive fields, each one interacting with a distinct region of the image. These “implicit patches” are not derived from explicitly partitioning the image but rather from fragmenting how the network perceives the input. This prevents overlap of receptive fields and avoids creating a “bottleneck” of information among neurons. Each neuron in the final layer independently evaluates a specific portion of the image, and the final output is obtained by averaging these individual evaluations (\hyperref[fig:figure5]{Fig. \ref{fig:figure5}}).

This mechanism has been successfully adopted in multiple subsequent style transfer models after Pix2Pix, such as CycleGAN \cite{key2} and CUT \cite{key3}, proving its effectiveness in diverse applications.

\textbf{Some Important Notes on PatchGANs}

\begin{itemize}
\item \textbf{Patch size and network depth:} In the initial layers of a convolutional network, neurons have relatively small receptive fields as they see only local regions of the image. As we add more layers, each neuron aggregates information from multiple neurons in the previous layer, thereby expanding its receptive field. Consequently, a deeper network will have larger output receptive fields. In a PatchGAN, increasing the number of layers inevitably enlarges the patch size. However, the patch size is also influenced by the size of the convolution filters. \textbf{In our models, we vary network depth to change patch size: a deeper network produces larger patches, while a shallower network yields smaller patches} (\hyperref[fig:figure6]{Fig. \ref{fig:figure6}}).

\hypertarget{point2}{\item \textbf{Separating feature maps:} In a PatchGAN, it is possible to segment the feature maps according to the implicit patches corresponding to the neurons in the final layer. Since each neuron in this layer is linked to a specific region of the image, the feature maps produced throughout the network can be partitioned into as many parts as there are final-layer neurons. Each portion of these maps corresponds to a precise region of the image, thus forming a bridge between that region and its associated neurons in the final layer.}
\end{itemize}

As indicated in \hyperlink{point2}{point 2}, PatchGAN provides a fragmented view of an image’s feature maps. This approach can address the spatial coherence loss of style by, for instance, computing Gram matrices for each patch and using an attention mechanism to optimally weight them.

\subsubsection{The CUT Model and the Work of Park, Taesung et al.\ \cite{key3}}

In \cite{key3}, Taesung Park introduces contrastive learning on “implicit” patches (PatchGAN) using InfoNCE \cite{key3}. The idea is to associate patches of the input image with patches of the generated image by maximizing their similarity in a learned feature space, thus preserving content while altering appearance. Improved correspondences are considered “positives,” while non-correspondences are considered “negatives.” This approach primarily aims to preserve content during style transfer.

\subsubsection{Diffusion Models and Style Transfer}

Diffusion models \cite{sohl2015deep}, \cite{ho2020denoisingdiffusionprobabilisticmodels} have recently proven to be effective at image generation. Several studies \cite{hamazaspyan2023diffusion, jeong2023training,wu2022unifying, yang2023zero, zhang2023inversion} explore their use in style transfer. These approaches manipulate the latent space of a pretrained diffusion model, such as “Stable Diffusion,” to separate style from content in an image.

However, applying these methods to style or weather classification would require a sufficiently general and powerful diffusion model, demanding substantial computational resources, which makes real-time, large-scale use less feasible.

Nonetheless, the capacity of diffusion models to separate style and content is highly promising. Through their progressive denoising process, they capture fine details of an image, including those related to style. These methods are thus beyond the scope of this paper.

\subsection{Review of Weather Detection Methods}\label{sec4}

As mentioned in the introduction, we will compare approaches derived from style transfer algorithms with standard weather detection methods. This section provides an overview of these latter methods. In recent years, numerous studies on weather classification have been published \cite{introvigne2024real, xia2020resnet15, naufal2022weather, papadimitriou2023advancing, patel2021weather, mittal2023classifying, kukreja2023multi, ibrahim2019weathernet, guerra2018weather, li2023study, rani2023weather, pikun2022image, dahmane2020analyse}. We have divided them into two categories: those relying on “classic” convolutional neural networks and those integrating attention mechanisms.

\subsubsection{Weather Classification with Convolutional Neural Networks}

Among the works using convolutional networks for weather classification \cite{introvigne2024real, xia2020resnet15, naufal2022weather, papadimitriou2023advancing, patel2021weather, mittal2023classifying, kukreja2023multi, ibrahim2019weathernet, guerra2018weather}, Jingming Xia’s work \cite{xia2020resnet15} stands out. The author proposes truncating ResNet50 to produce a lighter model called ResNet15, reduced to 15 layers. This simplification improves performance from 85.76\% to 96.03\%. Although the main motivation is to reduce complexity and speed up computations, these results raise the question of whether very deep networks are truly necessary for weather detection, especially given Gatys’ work \cite{gatys2016image} indicating that higher layers of detection models such as VGG \cite{simonyan2015deepconvolutionalnetworkslargescale} tend to focus on content over appearance.

\subsubsection{Weather Classification with Attention Mechanisms}

Among these studies \cite{li2023study, rani2023weather, pikun2022image} stands out by combining two types of attention: Convolutional Self-Attention (CSA) for capturing low-level features, and Atrous Spatial Attention (ASA) for extracting high-level features. This method enables the model to better represent various scales of features in weather images by leveraging lower layers for local information and higher layers, with larger receptive fields, for global information.

The author achieves 97.47\% accuracy on the MWD (Multi-class Weather Database) \cite{lin2017rscm} using this approach. Our own results obtained with shallower networks invite a reevaluation of whether very deep networks are required to capture stylistic information or, in our case, to detect weather.

\section{Presentation of Our Models}\label{sec5}

\begin{table}[!t]
    \centering
    \renewcommand{\arraystretch}{1.5}
    \caption{Machine Specifications}
    \label{tab:tableau1}
    \resizebox{0.95\columnwidth}{!}{
    \begin{tabular}{|c|c|c|}
    \hline
    \rowcolor{gray!30}
    \textbf{RAM} & \textbf{CPU} & \textbf{GPU}  \\
    \hline
    32.0 GB & Intel® Core™ i9-12900F 12th Gen × 24 & NVIDIA RTX A4500  \\
    \hline
    \end{tabular}}
\end{table}

The design of the models presented in this section primarily relies on the style transfer works of \hyperlink{Gatys}{Leon A. Gatys et al.} \cite{gatys2016image} and \hyperlink{Isola}{Phillip Isola et al.} \cite{key1}. The first two models draw upon Gatys’ work \cite{gatys2016image}, while the last model is based on Isola’s approach.

Analysis of Leon A. Gatys’ work \cite{gatys2016image} shows that a deep model pre-trained for object classification is not necessarily optimal for style or weather classification. The higher layers of such models are designed to ignore variations in brightness, contrast, and color, which are crucial for distinguishing styles and, under our hypothesis, for distinguishing weather conditions.

This led us to truncate the higher layers of a network such as ResNet50 (see Fig.~\ref{fig:figure7}), using pre-trained weights, in order to improve performance. This approach proved effective in the work of Jingming Xia \cite{xia2020resnet15}, where the F1 score of ResNet50 increased from 85.76\% to 96.03\% by reducing the model to ResNet15. We also explore the use of the Gram matrix, in the spirit of Gatys, to capture stylistic elements.

\subsection{Model 1: Truncated ResNet50 with MoCoV3 Weights}

\begin{figure}[!t]
    \centering
    \includegraphics[width=0.9\columnwidth]{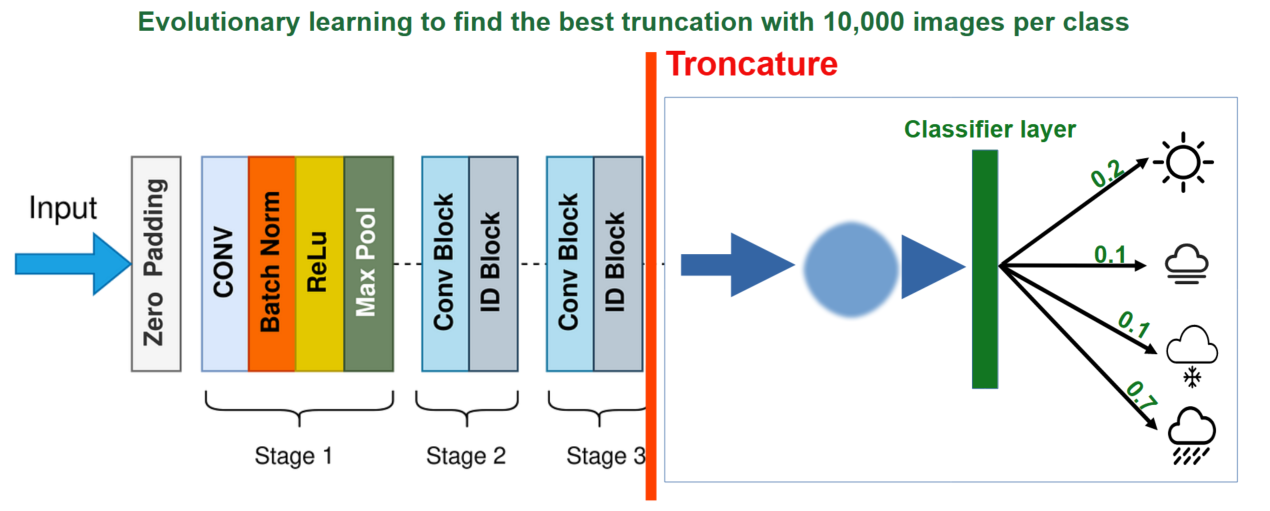}
    \caption{Retraining the “Truncated ResNet50” model with MoCoV3 weights.}
    \label{fig:figure7}
\end{figure}

The “Truncated ResNet50” model is a simplified version of ResNet50, using pre-trained MoCoV3 \cite{chen2021empiricalstudytrainingselfsupervised} weights known for producing high-quality features. Unlike Gatys, we retrain this truncated version. The truncation is carried out via an evolutionary algorithm that generates multiple configurations with different depths (e.g., 9, 13, or 20 layers). The embeddings from these versions feed a fully connected classifier with four outputs (see Fig.~\ref{fig:figure7}). The search for the optimal configuration (number of layers and hyperparameters) relies on a set of 40,000 images evenly distributed among the classes fog, rain, snow, and sun.

\textbf{Once a high-performing model is identified, it is retrained using the entire set of 120,000 data points (30,000 per class: fog, rain, snow, sun).}

\subsection{Model 2: Truncated ResNet50 with MoCoV3 Weights + Gram Matrix + Attention}

\begin{figure}[!t]
    \centering
    \includegraphics[width=0.9\columnwidth]{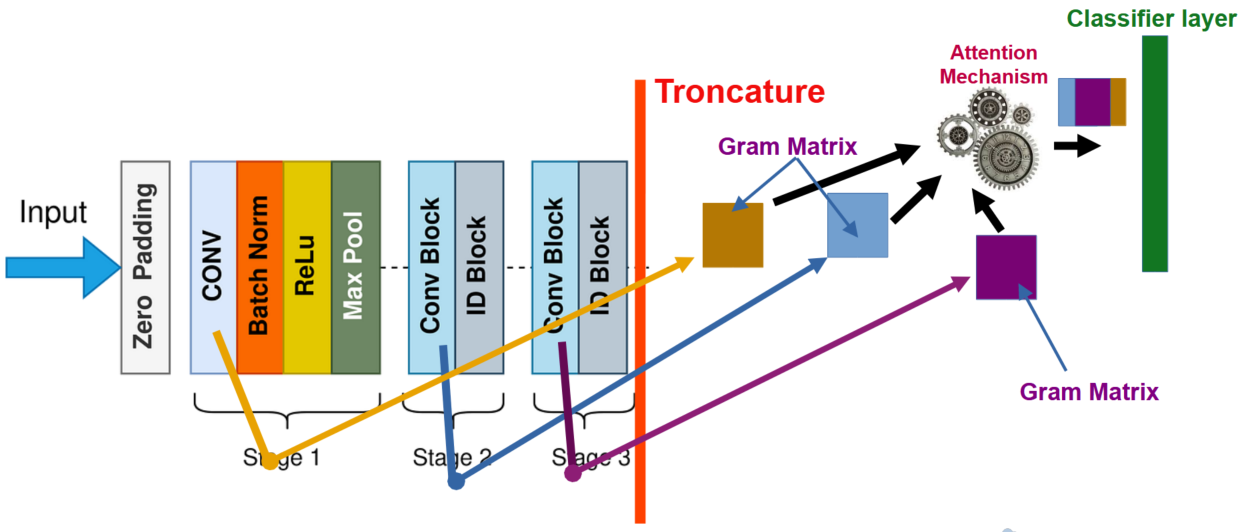}
    \caption{"Truncated ResNet50 + Gram Matrix + Attention": computing the Gram matrix for each layer of the truncated model, followed by an attention mechanism to weight the layers.}
    \label{fig:figure8}
\end{figure}

Training remains similar to the previous model, but here we compute the Gram matrix for every layer of the truncated model. An attention mechanism is used to weight and aggregate these matrices, thereby incorporating the central observation of Gatys’ work \cite{gatys2016image}: depending on the layer, stylistic elements can be very fine or more global. The attention mechanism helps select the most relevant information for prediction (see Fig.~\ref{fig:figure8}). The best model obtained by this strategy is referred to as “Truncated ResNet50 + Gram Matrix + Attention.”

\subsection{Multi-Patch Simple PatchGAN}

\begin{figure}[!t]
    \centering
    \includegraphics[width=0.9\columnwidth]{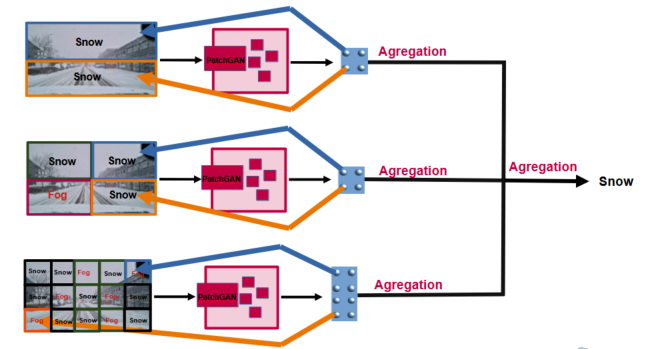}
    \caption{Combining several PatchGANs \cite{key1} with different patch sizes for improved robustness: Multi-Patch PatchGAN structure}
    \label{fig:figure9}
\end{figure}

We propose the “Multi-Patch Simple PatchGAN” model, which combines several PatchGANs \cite{key1} using three patch sizes: small, medium, and large (see Fig.~\ref{fig:figure9}). This approach allows for analyzing stylistic features at various scales. The optimal patch sizes are determined via an evolutionary algorithm that produces a population of models with different configurations.

We have introduced three weather detection models inspired by style transfer research, based on different logical approaches. The following section describes the dataset used for training and testing, and Section~V presents the obtained results.

\section{Datasets}\label{sec4}

Our data primarily come from two main sources:

\begin{itemize}
    \item Public databases or those from previous work in weather classification, such as \href{https://www.kaggle.com/datasets/abdulkarimkhalid/foggy-heavy-rain-and-sunny-images}{\textit{Kaggle (click to see)}}, \href{https://www.kaggle.com/datasets/vijaygiitk/multiclass-weather-dataset}{\textit{Kaggle850}}, \href{https://mwidataset.weebly.com/}{\textit{MWI (1996 images)}} \cite{zhang2015multi}, and Image2Weather \cite{chu2016image2weather}, for a total of 188,636 images.
    \item A dataset compiled by the Cerema I.T.S. research team, consisting of images and videos obtained from the Internet: 132,000 images (see Fig.~\ref{fig:figure8}). Available here: \href{https://drive.google.com/drive/folders/1eqnTRWLPH1FbhZdvnazt01fxp0vUN47n?usp=sharing}{\textbf{\textit{dataset}}} or by \href{https://github.com/Hamedkiri/heuristique_style_transfer_code.git}{\textbf{\textit{Github}}}.
\end{itemize}

For training, we exclusively used the Cerema dataset, comprising 132,000 images divided into four classes (sun, fog, rain, snow), with 33,000 images per class. We reserved 120,000 images (30,000 per class) for training and validation, and 12,000 images (3,000 per class) for testing. The other databases were used to test the generalization ability of our models, since they were not used for training and differ significantly from the Cerema dataset.

\begin{figure}[!t]
    \centering
    \includegraphics[width=0.9\columnwidth]{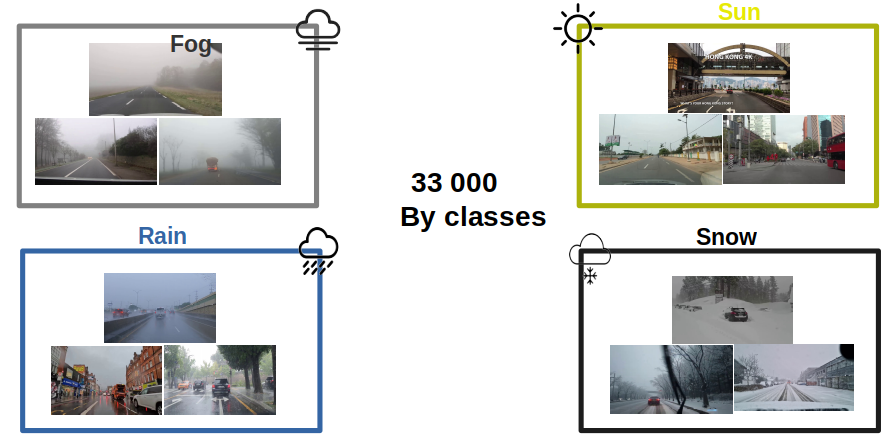}
    \caption{Examples of images from the Cerema dataset.}
    \label{fig:figure50}
\end{figure}

\section{Results}\label{sec4}

\subsection{Comparison with Similar Works}

\begin{figure}[!t]
    \centering
    \includegraphics[width=0.6\linewidth]{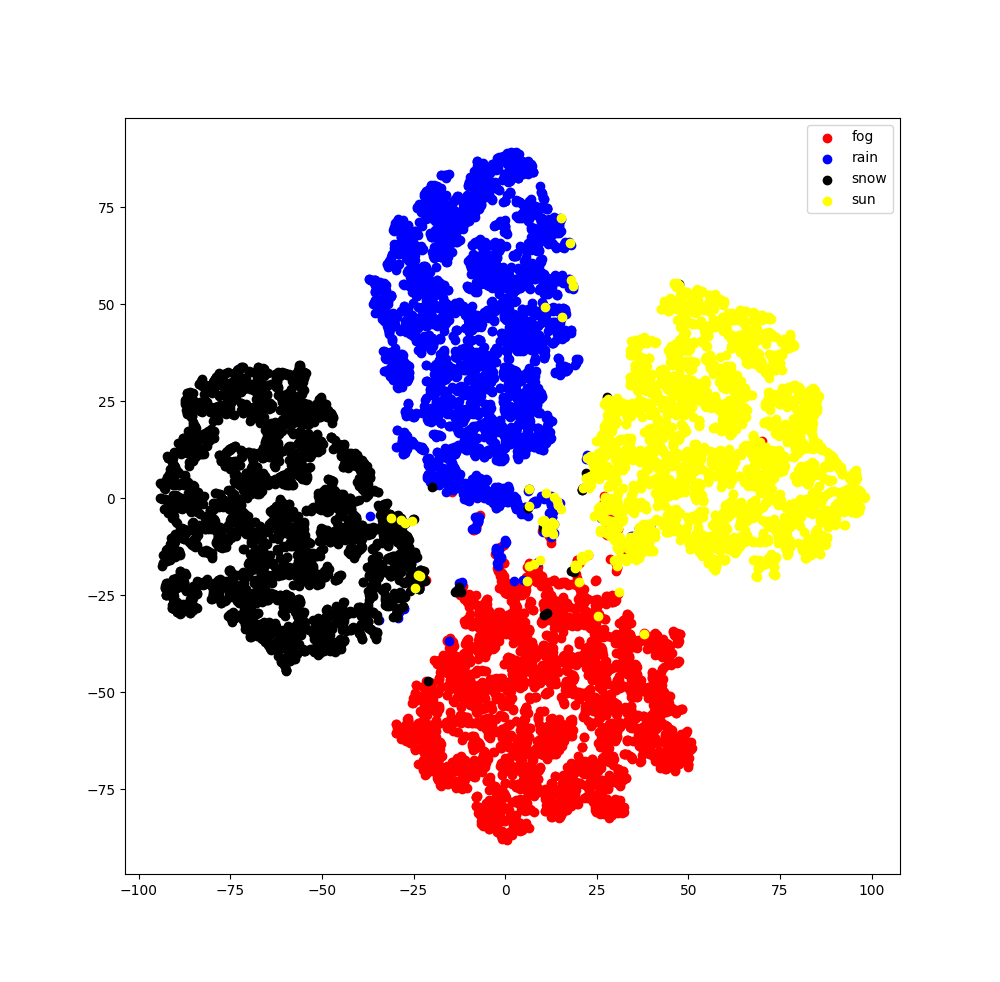}
    \caption{t-SNE visualization of “Truncated ResNet50” embeddings}
    \label{fig:figure11}
    
    \vspace{0.5cm}
    
    \includegraphics[width=0.6\linewidth]{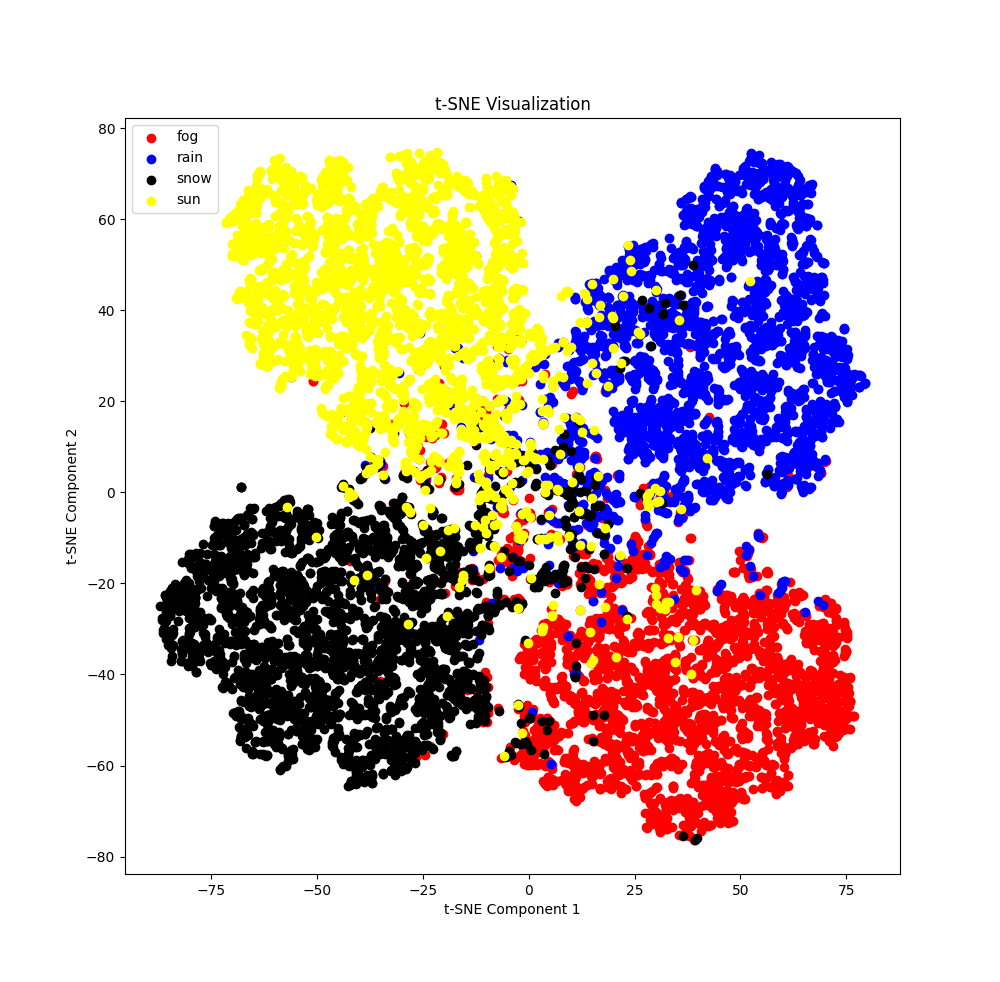}
    \caption{t-SNE visualization of “Multi-Patch Simple PatchGAN” embeddings}
    \label{fig:figure12}
    
    \vspace{0.5cm}
    
    \includegraphics[width=0.6\linewidth]{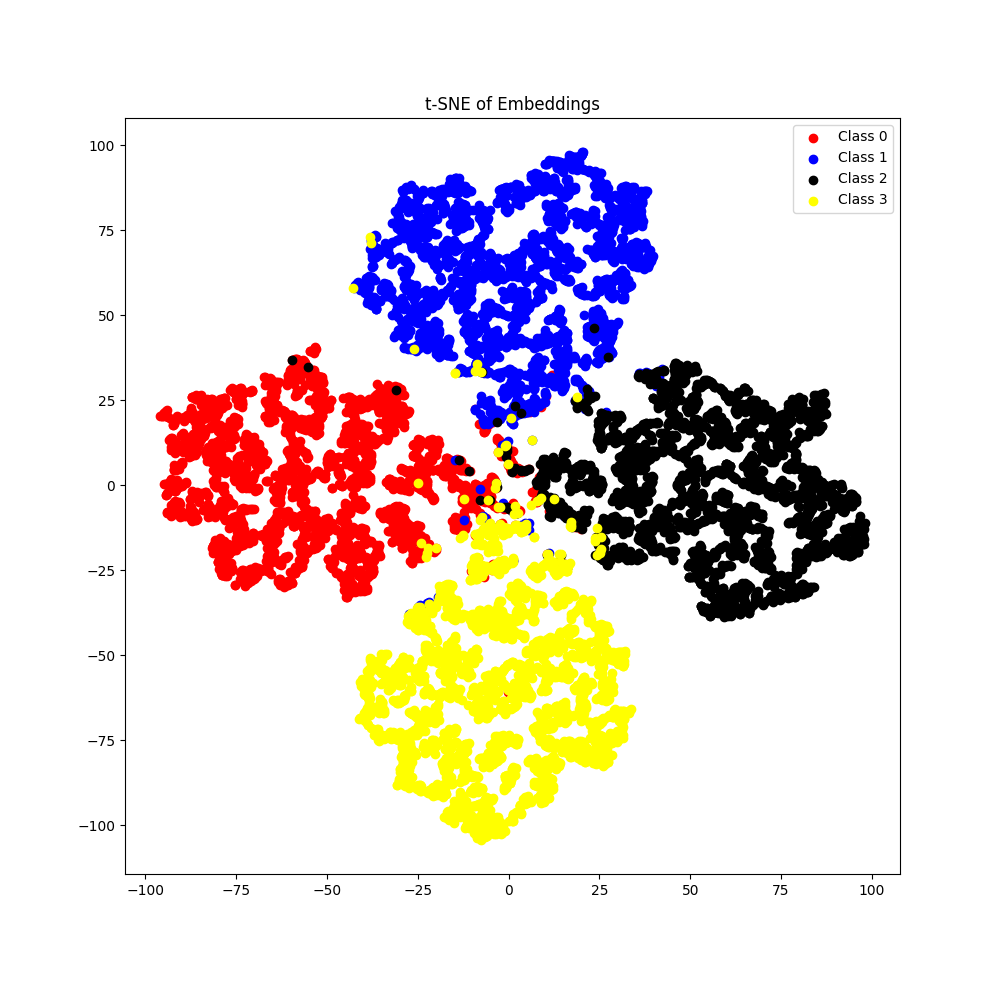}
    \caption{t-SNE visualization of “Truncated ResNet50 + Gram Matrix + Attention” embeddings}
    \label{fig:figure13}
\end{figure}

\hypertarget{results}{We} compared the performance of our models against 13 similar studies \cite{introvigne2024real, xia2020resnet15, naufal2022weather, papadimitriou2023advancing, patel2021weather, mittal2023classifying, kukreja2023multi, ibrahim2019weathernet, guerra2018weather, li2023study, rani2023weather, pikun2022image, dahmane2020analyse}. Table~\ref{tab:tableau2} summarizes the performance of our models as well as two reference studies \cite{li2023study,xia2020resnet15}. Our models match or surpass the state-of-the-art on comparable datasets. In particular, “Truncated ResNet50” (precision: 98.84\%, recall: 98.84\%, F1-score: 98.84\%) and “Truncated ResNet50 + Gram Matrix + Attention” (precision: 98.30\%, recall: 98.30\%, F1-score: 98.29\%) achieve excellent results on a test set of 12,000 images (3,000 per class) while running in real time on a GPU (see Table~\ref{tab:tableau1}). The t-SNE((t-Distributed Stochastic Neighbor Embedding) visualizations of our models’ embeddings, before the fully connected classification layer, are shown in Fig.~\ref{fig:figure11}, Fig.~\ref{fig:figure12}, and Fig.~\ref{fig:figure13}, revealing a clear separation among image classes.

\begin{table}[!t]
    \centering
    \renewcommand{\arraystretch}{1.5}
    \caption{Comparison of three models based on style transfer with two other "traditional" methods (F1 Score)}
    \label{tab:tableau2}
    \resizebox{\columnwidth}{!}{
    \begin{tabular}{|p{3cm}|p{2.5cm}|p{2cm}|p{2cm}|p{2.5cm}|p{3cm}|}
    \hline
    \rowcolor{gray!30}
    \textbf{Model Name} & \textbf{Test Set} & \textbf{F1 Score} & \textbf{Training Epochs} & \textbf{Execution} & \textbf{Number of Parameters}\\
    \hline
    \multicolumn{6}{|c|}{\cellcolor{gray!50}\textbf{Other Works}} \\
    \hline
    VIT + Enhanced Attention Module~\cite{li2023study} & MWD + WEAPD (6688) & 92.73\% & Unknown & Unknown (not real-time) & Unknown\\
    \hline
    ResNet15~\cite{xia2020resnet15} & Weather Dataset (983) & 96.03\% & Unknown & 4.4 I/s (CPU) and 163.3 I/s (GPU) & 5,000,000 \\
    \hline
    \multicolumn{6}{|c|}{\cellcolor{gray!50}\textbf{Our Models}} \\
    \hline
    \multicolumn{6}{|l|}{\cellcolor{gray!20}\textbf{Group 1: Truncated ResNet50}} \\
    \hline
    Truncated ResNet50 & Cerema Dataset (12,000) & \textbf{98.84\%} & 10 & 200 I/s & 24,033,604 \\ 
    \hline
    Truncated ResNet50 + Gram Matrix + Attention & Cerema Dataset (12,000) & 98.29\% & 50 & 143.51 I/s & 39,926,736 \\
    \hline
    \multicolumn{6}{|l|}{\cellcolor{gray!20}\textbf{Group 2: PatchGAN}} \\
    \hline
    Multi-Patch Simple PatchGAN & Cerema Dataset (12,000) & 93.88\% & 500 & 71.82 I/s & 28,896,396 \\
    \hline
    \end{tabular}
    }
\end{table}

\subsection{Generalization Capability of the Models}

We evaluate here the performance of our models on datasets not used during training, in order to measure their generalization capability. Overall, most of our models generalize well. The best model, “Truncated ResNet50,” maintains recall and F1 scores of no less than 74.82\% and 79.09\%, respectively (Table~\ref{tab:tableau3}).

\begin{table}[!t]
    \centering
    \renewcommand{\arraystretch}{1.5}
    \caption{Ability to generalize our models via tests on databases that did not participate in training (F1 Score)}
    \label{tab:tableau3}
    \resizebox{\columnwidth}{!}{
    \begin{tabular}{|p{3cm}|p{2cm}|p{2cm}|p{2cm}|p{2cm}|p{2cm}|}
    \hline
    \rowcolor{gray!30}
    \textbf{Model Name} & \textbf{Cerema (12,000)} & \textbf{MWI (1,996)} & \textbf{Image2Weather (2,000)} & \textbf{Kaggle (2,000)} & \textbf{Kaggle (850)}  \\
    \hline
    \multicolumn{6}{|l|}{\cellcolor{gray!20}\textbf{Group 1: Truncated ResNet50}} \\
    \hline
    Truncated ResNet50 & 98.84\% & 79.39\% & 82.35\% & 94.29\% & 79.06\% \\
    \hline
    Truncated ResNet50 + Gram Matrix + Attention & 98.29\% & 78.49\% & 81.33\% & 94.18\% & 76.74\% \\
    \hline
    \multicolumn{6}{|l|}{\cellcolor{gray!20}\textbf{Group 2: PatchGAN}} \\
    \hline
    Multi-Patch Simple PatchGAN & 93.88\% & 70.19\% & 74.79\% & 88.93\% & 65.01\% \\
    \hline
    \end{tabular}}
\end{table}

\subsection{Interpretability and Real Deployment}

\begin{figure}[!t]
    \centering
    \begin{subfigure}[b]{0.48\columnwidth}
        \centering
        \includegraphics[width=\linewidth]{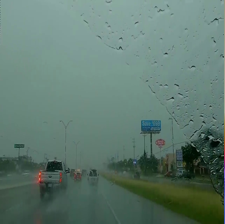}
        \caption{}
        \label{fig:figure14a}
    \end{subfigure}
    \hfill
    \begin{subfigure}[b]{0.48\columnwidth}
        \centering
        \includegraphics[width=\linewidth]{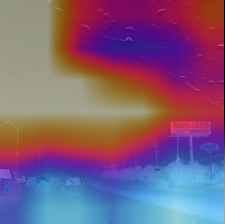}
        \caption{}
        \label{fig:figure14b}
    \end{subfigure}
    \caption{Grad-CAM on a rainy scene: (a) original image and (b) important areas highlighted and superimposed on the image. The brighter the color, the more the area plays an important role.}
    \label{fig:figure14}
\end{figure}

Since neural networks are often considered black boxes, Grad-CAM provides valuable interpretability (Fig.~\ref{fig:figure14}). It highlights the specific areas of an image on which the “Truncated ResNet50” model bases its predictions. In many instances, these areas coincide with what a human would also deem important, thereby reinforcing confidence in the model’s decisions (see Fig.~\ref{fig:figure14}). Moreover, developed within the framework of the \href{https://roadview-project.eu/}{\textit{Roadview}} European project, this model is intended for integration into autonomous vehicles, with real-world testing. All the models presented here are optimized for real-time operation, at speeds exceeding 20 images per second.

\section{Limitations of the Models}

\begin{figure}[!t]
    \centering
    \includegraphics[width=0.9\columnwidth]{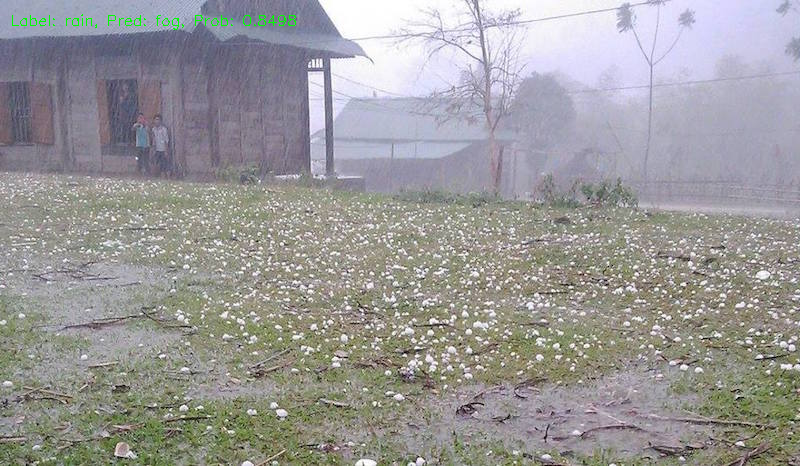}
    \caption{Simultaneous presence of rain and fog in one image, a subtlety that our "Truncated ResNet50" algorithm fails to capture.}
    \label{fig:figure80}
\end{figure}

Our tests have revealed certain inherent limitations in our models. First, they encounter difficulties when an image simultaneously exhibits two weather conditions, as illustrated in Fig.~\ref{fig:figure80}. Indeed, our models were trained to strictly distinguish among four weather conditions, which pushes them to make overly categorical, and sometimes unrealistic, classifications.

\begin{figure}[!t]
    \centering
    \includegraphics[width=0.9\columnwidth]{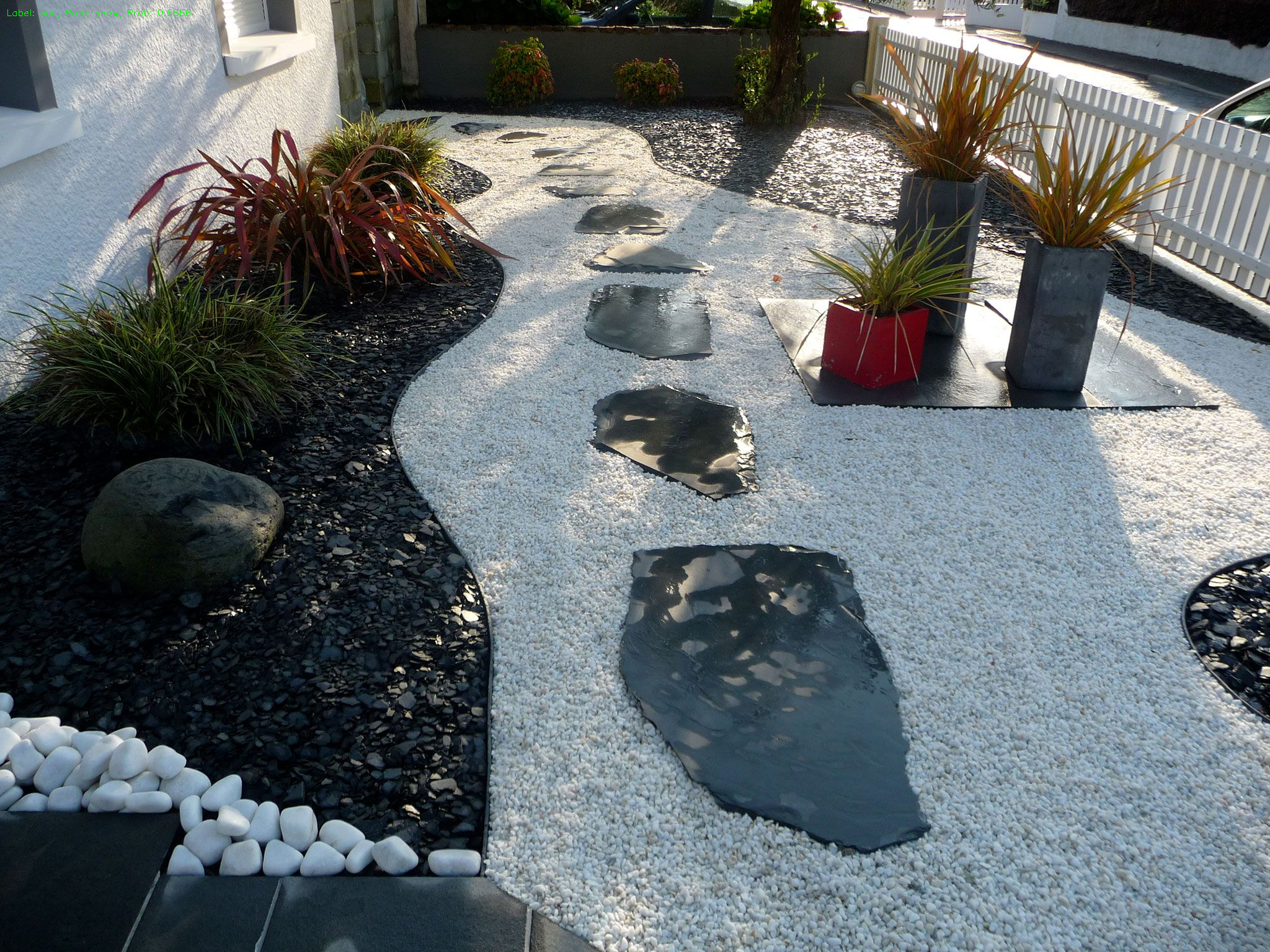}
    \caption{The "Truncated ResNet50" model mistakes gravel for snow in its prediction.}
    \label{fig:figure81}
\end{figure}

\begin{figure}[!t]
    \centering
    \begin{subfigure}[b]{0.48\columnwidth}
        \centering
        \includegraphics[width=\linewidth]{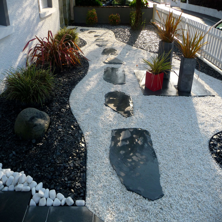}
        \caption{}
        \label{fig:figure14a}
    \end{subfigure}
    \hfill
    \begin{subfigure}[b]{0.48\columnwidth}
        \centering
        \includegraphics[width=\linewidth]{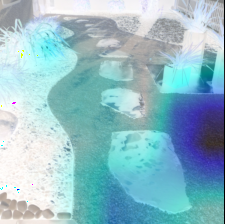}
        \caption{}
        \label{fig:figure14b}
    \end{subfigure}
    \caption{Grad-CAM shows that the model relies on the gravel for its prediction: (a) original image, (b) highlighted key areas. The brighter the color, the more the area plays an important role.}
    \label{fig:figure82}
\end{figure}

Furthermore, for the same reasons, the models often rely on visual elements that are not directly related to the weather conditions. For example, white gravel on the ground can be interpreted as snow, leading the algorithm to erroneously predict a snowy scene, as shown in Fig.~\ref{fig:figure81} and Fig.~\ref{fig:figure82}.

To overcome these limitations, we are working on a more sophisticated model designed to incorporate these subtleties and reduce classification errors.

\section{Application of Our Models to Other Tasks}

\subsection{Melanoma Detection}

\hypertarget{otherTasks}{The} “SIIM-ISIC 2020 Melanoma Classification Challenge” aimed to develop AI models to detect melanoma, a type of skin cancer that is hard to distinguish from benign lesions. Early detection improves survival rates, and a high-performance model can assist dermatologists and improve access to care. Here, the difference between classes is primarily based on the style or appearance of the lesions.

\begin{figure}[!t]
    \centering
    \begin{subfigure}[b]{0.48\columnwidth}
        \centering
        \includegraphics[width=\linewidth]{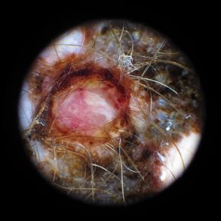}
        \caption{}
        \label{fig:figure15a}
    \end{subfigure}
    \hfill
    \begin{subfigure}[b]{0.48\columnwidth}
        \centering
        \includegraphics[width=\linewidth]{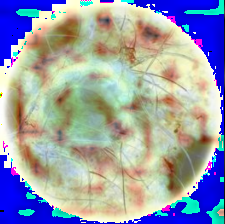}
        \caption{}
        \label{fig:figure15b}
    \end{subfigure}
    
    \vspace{0.5cm}

    \begin{subfigure}[b]{0.48\columnwidth}
        \centering
        \includegraphics[width=\linewidth]{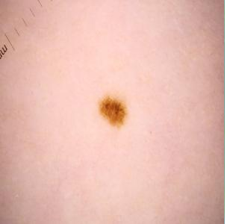}
        \caption{}
        \label{fig:figure15c}
    \end{subfigure}
    \hfill
    \begin{subfigure}[b]{0.48\columnwidth}
        \centering
        \includegraphics[width=\linewidth]{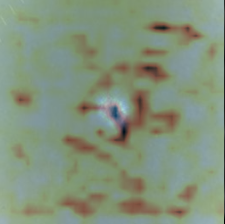}
        \caption{}
        \label{fig:figure15d}
    \end{subfigure}
    
    \caption{Grad-CAM of the “Truncated ResNet50” model on melanoma images: (a, b) malignant melanoma case and (c, d) benign case. On the right side, the heatmap is superimposed on the original image.}
    \label{fig:figure15}
\end{figure}

\begin{figure}[!t]
    \centering
    \includegraphics[width=0.9\columnwidth]{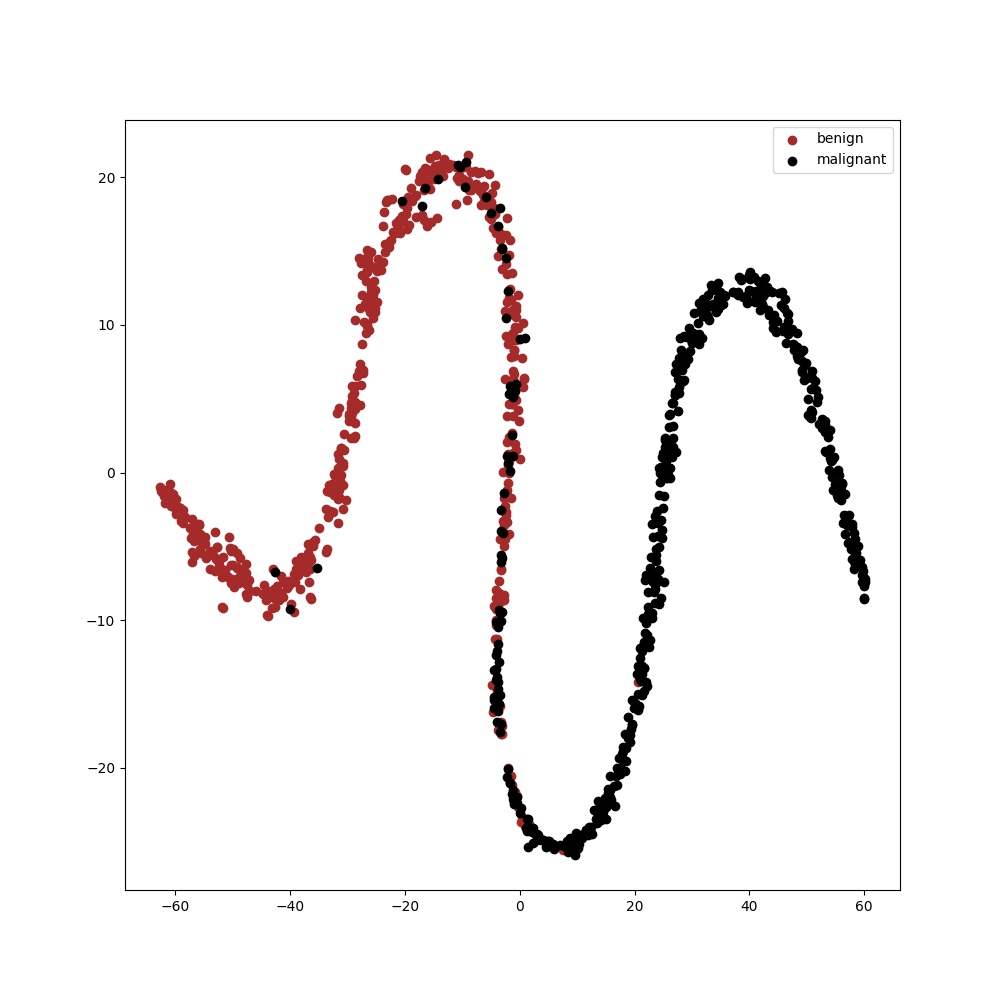}
    \caption{t-SNE visualization of the “Truncated ResNet50” model applied to melanoma images.}
    \label{fig:figure16}
\end{figure}

The dataset contained about 33,000 images (melanomas and benign lesions) and 10,982 unlabeled images for evaluation. The winning model \cite{ha2020identifyingmelanomaimagesusing}, based on EfficientNet, achieved an AUC of 0.9600 in cross-validation. Our “Truncated ResNet50” model, trained on the \href{https://www.kaggle.com/datasets/hasnainjaved/melanoma-skin-cancer-dataset-of-10000-images}{\textit{Melanoma Skin Cancer Kaggle Dataset}}, obtains an AUC of 0.9758 on a test set of 1,000 images. In Fig.~\ref{fig:figure15}, Grad-CAM highlights the zones of the image used by the model for prediction. Fig.~\ref{fig:figure16} shows the t-SNE projection, illustrating a clear distinction between benign and malignant embeddings.

Although our models were designed for weather detection, they also prove to be highly effective at classifying image style in general.

\section{Conclusion}

In this paper, we presented and validated neural network architectures designed for real-time weather classification from images, offering significant improvements in both accuracy and generalization. By leveraging style-transfer-based approaches, such as “Truncated ResNet50” and “Multi-Patch PatchGAN,” we achieve high F1 scores across diverse datasets. However, our models can sometimes be sensitive to elements unrelated to weather, such as a “white wall” mistakenly interpreted as snow.

To overcome these limitations, we are currently developing a multi-task model based on our hypothesis: “weather primarily affects the appearance of objects in an image.” By exploiting stylistic features, this model aims to go beyond merely classifying weather and also assess, for example, the intensity of the phenomenon, visibility, road and sky conditions, or the presence of glare. Such an approach paves the way for more robust and versatile systems.

\appendix

Our model codes are available here: \href{https://github.com/Hamedkiri/heuristique_style_transfer_code.git}{\textbf{\textit{code}}} and \href{https://drive.google.com/drive/folders/11Pllunglo-_XcZSI80WheTKOeqceW9II?usp=sharing}{\textbf{\textit{model weights}}}. The datasets used for training and testing the models are accessible here: \href{https://drive.google.com/drive/folders/1eqnTRWLPH1FbhZdvnazt01fxp0vUN47n?usp=sharing}{\textbf{\textit{datasets}}}.

\printbibliography[sorting=keysort]

\end{document}